\documentclass[11pt, logo, onecolumn, copyright, colorlinks=true, allcolors=blue]{nvidiatechreport}
\usepackage{graphicx} 
\usepackage{subcaption,ragged2e}
\usepackage[round]{natbib}

\usepackage[normalem]{ulem}
\usepackage[utf8]{inputenc} 
\usepackage[T1]{fontenc}    
\usepackage{url}
\usepackage{tikz}
\usepackage{enumitem}
\usetikzlibrary{positioning}
\usetikzlibrary{calc}
\usepackage{array}
\usepackage{booktabs}
\usepackage{wrapfig}
\usepackage{caption} 
\usepackage{listings}
\usepackage{adjustbox}
\usepackage{footnote}
\usepackage{multirow}
\usepackage{amsmath}
\usepackage{amsfonts}
\usepackage{breakcites}
\usepackage{blindtext}
\usepackage{svg}
\usepackage[most]{tcolorbox}
\usepackage{xcolor}
\usepackage{tabularx}
\usepackage{graphicx} 
\usepackage{makecell}
\newcommand{\newparagraph}[1]{\noindent\textbf{#1\hspace{0.5em}}}

\newtcolorbox{notebox}[1]{
  colback=white,
  colframe=green!50!black,
  coltitle=white,
  colbacktitle=green!50!black,
  title=#1,
  fonttitle=\bfseries,
  arc=6pt,
  boxrule=1pt,
  left=10pt,
  right=10pt,
  top=10pt,
  bottom=10pt,
}

\title{LatentMoE: Toward Optimal Accuracy per FLOP and Parameter in Mixture of Experts}
\author{\large Venmugil Elango, Nidhi Bhatia, Roger Waleffe, Rasoul Shafipour, Tomer Asida, Abhinav Khattar, Nave Assaf, Maximilian Golub, Joey Guman, Tiyasa Mitra, Ritchie Zhao, Ritika Borkar, Ran Zilberstein, Mostofa Patwary, Mohammad Shoeybi, Bita Rouhani}
\date{}

\begin{document}
\makeatletter
\DeclareRobustCommand\onedot{\futurelet\@let@token\@onedot}
\def\@onedot{\ifx\@let@token.\else.\null\fi\xspace}

\def\eg{\emph{e.g}\onedot} \def\Eg{\emph{E.g}\onedot}
\def\ie{\emph{i.e}\onedot} \def\Ie{\emph{I.e}\onedot}
\def\cf{\emph{c.f}\onedot} \def\Cf{\emph{C.f}\onedot}
\def\etc{\emph{etc}\onedot} \def\vs{\emph{vs}\onedot}
\def\wrt{w.r.t\onedot} \def\dof{d.o.f\onedot}
\def\etal{\emph{et al}\onedot}
\makeatother

\newcommand{\todo}[1]{\textcolor{red}{TODO: #1}}
\newcommand{\R}{\mathbb{R}}
\newcommand{\act}{\operatorname{ACT}}
\newcommand{\topk}{\operatorname{TopK}}
\newcommand{\softmax}{\operatorname{Softmax}}
\newcommand{\moe}{\operatorname{MoE}}
\newcommand{\latentmoe}{\operatorname{\ell-MoE}}
\newcommand{\lmoeeff}{\latentmoe_\mathrm{eff}}
\newcommand{\lmoeacc}{\latentmoe_\mathrm{acc}}
\newcommand{\lmoeefflarge}{\latentmoe_\mathrm{eff-12B}}
\newcommand{\wfconei}{\ensuremath{W_{\text{FC1}}^{(i)}}}
\newcommand{\wfctwoi}{\ensuremath{W_{\text{FC2}}^{(i)}}}
\newcommand{\wgatei}{\ensuremath{W_{\text{gate}}^{(i)}}}
\newcommand{\nvlinkbw}{\mathrm{BW}_\mathrm{NVL}}
\newcommand{\hbmbw}{\mathrm{BW}_\mathrm{HBM}}
\newcommand{\ep}{\mathrm{EP}}
\newcommand{\ttotal}{t_\mathrm{total}}
\newcommand{\texp}{t_\mathrm{exp}}

\begin{abstract}
\large \textbf{Abstract.}
Mixture of Experts (MoEs) have become a central component of many state-of-the-art open-source and proprietary large language models.
Despite their widespread adoption, it remains unclear how close existing MoE architectures are to optimal with respect to inference cost, as measured by accuracy per floating-point operation and per parameter. In this work, we revisit MoE design from a hardware-software co-design perspective, grounded in empirical and theoretical considerations. We characterize key performance bottlenecks across diverse deployment regimes, spanning offline high-throughput execution and online, latency-critical inference. Guided by these insights, we introduce \textbf{LatentMoE}, a new model architecture resulting from systematic design exploration and optimized for maximal accuracy per unit of compute. Empirical design space exploration at scales of up to 95B parameters and over a 1T-token training horizon, together with supporting theoretical analysis, shows that LatentMoE consistently outperforms standard MoE architectures in terms of accuracy per FLOP and per parameter. Given its strong performance, the LatentMoE architecture has been adopted by the flagship Nemotron-3 Super and Ultra models and scaled to substantially larger regimes, including longer token horizons and larger model sizes, as reported in~\cite{nvidia2025nvidianemotron3efficient}.

\end{abstract}

\maketitle

\section{Introduction}
Transformer-based large language models underpin a wide range of modern AI systems, from conversational agents to code generation and scientific reasoning. As these models continue to scale, practical deployment is increasingly constrained by inference cost, encompassing both computation and memory. As a result, a central objective in modern model design is to maximize achievable accuracy under fixed inference cost constraints.

Mixture-of-Experts (MoE) architectures have emerged as a promising approach towards this goal, enabling models to scale parameter count while keeping the number of Floating-point Operations (FLOPs) per token fixed. Despite their empirical success, the MoE design space remains poorly understood. Existing MoE architectures are largely motivated by high-level sparsity arguments and are optimized primarily for offline, throughput-oriented settings, with limited consideration of online deployments that impose strict latency, memory bandwidth, and communication constraints.

We argue that effective MoE design must be evaluated along two complementary dimensions: accuracy per FLOP and accuracy per parameter. While accuracy per FLOP captures computational efficiency, accuracy per parameter reflects memory footprint, memory bandwidth demands, routing-induced communication, and sharding overheads (factors that are often the dominant bottlenecks in interactive, low-latency inference). Neglecting these factors can lead to architectures that appear efficient in aggregate compute, yet incur substantial inefficiencies in practical deployment.


\begin{figure}[h!tb]
    \centering
    \begin{subfigure}[b]{0.42\textwidth}
        \includegraphics[width=\textwidth]{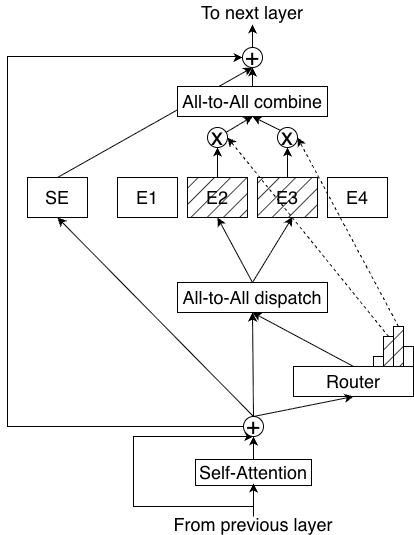}
        \caption{Standard MoE architecture.}
        \label{fig:standard_moe}
    \end{subfigure}
    \hfill
    \begin{subfigure}[b]{0.45\textwidth}
        \includegraphics[width=\textwidth]{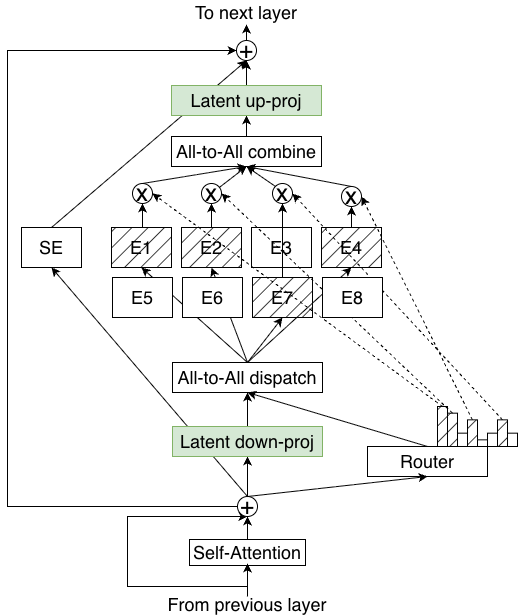}
        \caption{LatentMoE architecture.}
        \label{fig:latent_moe}
    \end{subfigure}
    \caption{Standard MoE vs. LatentMoE architectures. In LatentMoE, tokens are projected from the model hidden dimension $d$ into a smaller latent dimension $\ell$ for expert routing and computation, which reduces routed parameter loads and all‑to‑all traffic by a factor of $d / \ell$. We use this efficiency to increase the total number of experts and the top-k active experts per token by the same factor $d/\ell$, which improves the accuracy of the model while keeping overall inference cost approximately constant.}
    \label{fig:intro_fig}
\end{figure}

In this work, we revisit MoE architecture design from a hardware–software co-design perspective. Through a systematic analysis of existing MoE systems across the throughput–latency Pareto frontier, we identify key structural bottlenecks arising from expert parameterization, routing-induced all-to-all communication, and memory access patterns. Combined with detailed accuracy measurements and theoretical analysis, our study identifies structural inefficiencies in prevailing MoE designs that limit achievable accuracy per unit of inference cost.

Guided by these insights, we introduce LatentMoE, a new mixture-of-experts architecture explicitly optimized for both accuracy per FLOP and accuracy per parameter. LatentMoE decouples expert routing and computation from the model hidden dimension by projecting incoming activations into a shared low-dimensional latent space prior to expert processing (Figure~\ref{fig:intro_fig}). The latent dimension serves as a direct control knob for computational cost, communication volume, and expert parameter size. At iso-FLOP and iso-parameter count, projecting incoming activations into a lower-dimensional latent space enables a proportional increase in both the number of experts and the routing top-k, while maintaining constant inference cost. 

As we show both theoretically and empirically, simultaneously increasing expert count and combinatorial sparsity diversity improves the effective expressivity of the model, leading to higher achievable accuracy. Crucially, these gains arise without increasing memory bandwidth demands or communication overheads, making LatentMoE well suited for both latency-critical and throughput-oriented deployments. We validate the LatentMoE concept through pretraining experiments at scales of up to 95B parameters and over 1T tokens. Across all evaluated regimes, LatentMoE consistently improves upon standard MoE architectures, achieving higher accuracy at fixed inference cost or substantially lower inference cost at fixed accuracy. Given its strong performance, the LatentMoE architecture has been adopted by the flagship Nemotron-3 Super and Ultra models and scaled to substantially larger regimes, including longer token horizons and larger model sizes~\citep{nvidia2025nvidianemotron3efficient}.

\section{LatentMoE Core Design Principles}\label{sec:design_choices}
 
Before delving into the specifics of LatentMoE, we first take a systems-level view of what is required to deploy an MoE model that is both accurate and cost-efficient.

Throughout this section, we use Qwen3-235B-A22B as a running example for our modeling, with 
$N = 128$ experts, $K=8$ active experts per token, a hidden dimension $d=4096$, and an intermediate feed-forward dimension $m=1536$. For concreteness, we consider deployment on NVIDIA GB200 GPUs interconnected by a high-bandwidth NVLink fabric, which provides approximately $1800$~GB/s of bidirectional bandwidth per GPU (i.e., $\nvlinkbw=900$~GB/s per direction). To ensure that expert communication remains within a single NVLink domain, experts are distributed via expert parallelism across $\ep=64$ GPUs. Attention layers are executed using data parallelism over the same group of GPUs. Each GB200 GPU delivers a peak FP4 Tensor Core throughput of $F=10$ PFLOPs and an HBM memory bandwidth of $\hbmbw=8$~TB/s \citep{nvidia_blackwell}.

\subsection{Memory Bandwidth Bottleneck}
\begin{figure}[ht!]
    \centering
    \includegraphics[width=0.7\linewidth]{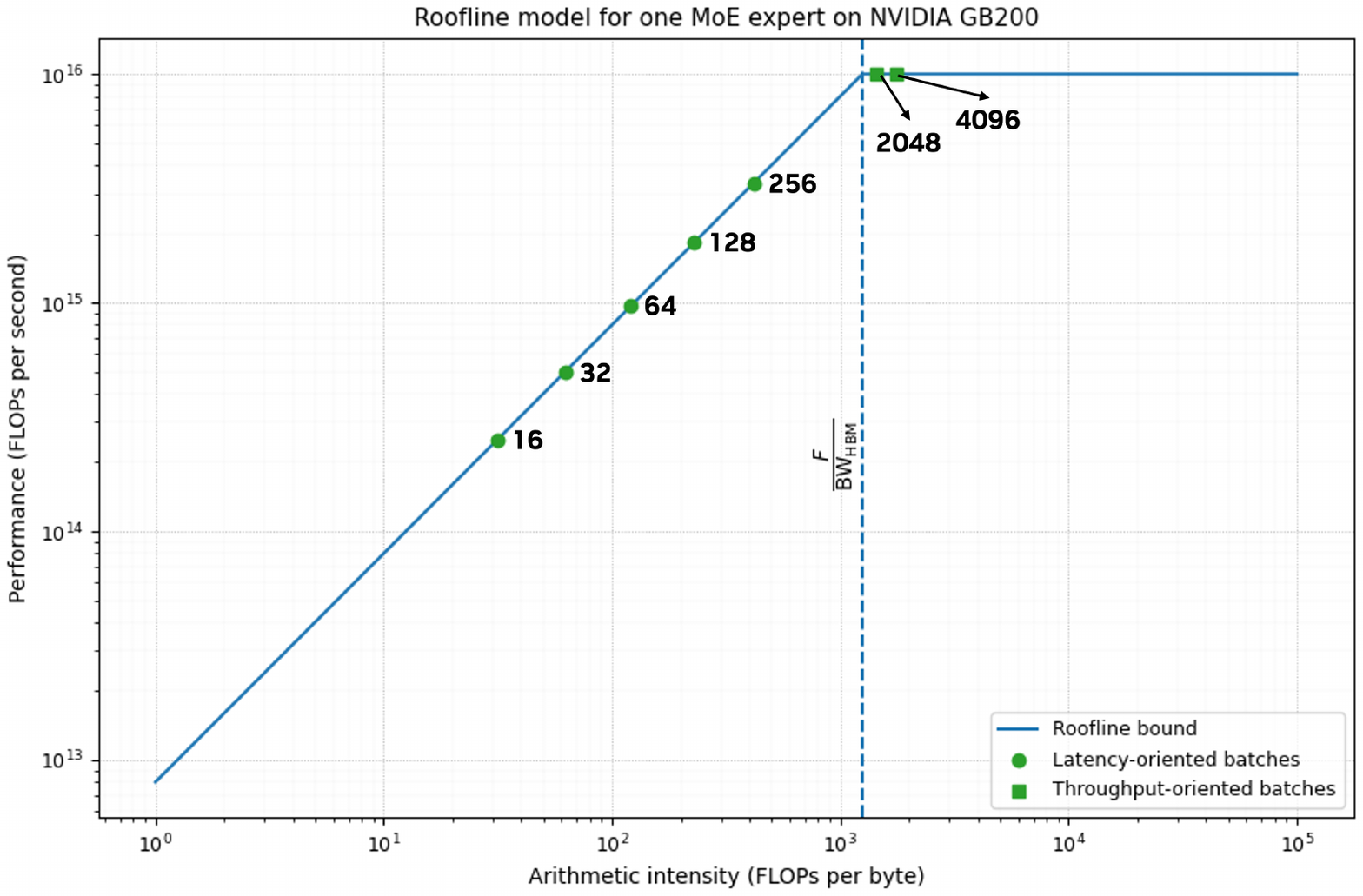}
    \caption{\textbf{Roofline Analysis for serving Qwen3-235B-A22B.}
    Operating points correspond to different per-expert token counts $\texp$ (i.e., effective expert batch sizes after MoE routing), mapped to arithmetic intensity $I=\frac{2 \cdot \texp \cdot d \cdot m}{d \cdot m + \texp \cdot (d+m)}$. At latency-critical batch sizes (low $I$), MoE expert computation is constrained by HBM bandwidth rather than compute, and the operating points lie in the bandwidth-bound regime.}
    \label{fig:roofline}
\end{figure}

In highly interactive (\ie, low-latency) settings that typically use small batch sizes, MoE computation is primarily bottlenecked by memory bandwidth. Figure~\ref{fig:roofline} provides a high-level roof-line analysis of performance versus arithmetic intensity.

For a GB200 system, a computation becomes compute-bound only if its arithmetic intensity (i.e., FLOPs per byte) exceeds:
\begin{equation*}
\frac{F}{\hbmbw} = \frac{10 \times 10^{15}}{8 \times 10^{12}} = 1250 \text{ FLOPs/byte}.
\end{equation*}

Let $\ttotal$ be the total number of tokens across the $\ep$ GPUs prior to MoE routing. Assuming a uniform distribution of tokens across experts, the number of tokens assigned to a single expert is: $\texp := \frac{\ttotal \cdot K}{N}$. In the Qwen3-235B-A22B example, with $N = 128$ and $\ep = 64$, each GPU hosts $N / \ep = 2$ experts; thus each GPU processes approximately $2 \cdot \texp$ expert tokens per MoE layer.

The FP4 compute cost for a single expert is $C_{\mathrm{exp}} = 2 \cdot \texp \cdot d \cdot m$, and the corresponding memory traffic in FP4 precision---accounting for weights, inputs, and intermediate activations---is given by $M_{\mathrm{exp}} = d \cdot m + \texp \cdot (d + m)$. Since each GPU processes two experts in our example, the arithmetic intensity $I$ is given by the ratio of the total compute to the total memory traffic:
\begin{equation*}
I = \frac{2 \cdot C_{\mathrm{exp}}}{2 \cdot M_{\mathrm{exp}}} = \frac{2 \cdot \texp \cdot d \cdot m}{d \cdot m + \texp \cdot (d + m)}.
\end{equation*}
To operate in the compute-bound regime, we require $I \ge 1250$. Substituting the Qwen3-235B-A22B parameters yields the condition:
\begin{equation*}
\frac{2 \cdot \texp \cdot d \cdot m}{d \cdot m + \texp \cdot (d + m)} \ge 1250 \quad \implies \quad \texp \ge 1418.
\end{equation*}
In typical latency-critical deployments, effective batch sizes are small, resulting in $\texp$ being on the order of a few hundred tokens—well below the threshold of 1418. Consequently, MoE experts operate in the memory-bound region of the roofline curve (Figure~\ref{fig:roofline}), where performance is limited by weight loading rather than compute capacity.

\begin{notebox}{Design Principle I}
In low-latency serving scenarios, MoE inference is typically dominated by the memory-bandwidth cost of loading model weights. Consequently, maximizing accuracy per parameter is critical for applications with high interactivity requirements.
\end{notebox}

\subsection{Communication Bottleneck}
In throughput-oriented settings, once experts become compute-bound, communication emerges as a significant contributor to end-to-end execution time in distributed settings. Expert parallelism mandates all-to-all token routing across devices, imposing an overhead that can control end-to-end execution time.
Recall that in our example configuration, each GPU hosts two experts. Since $\texp$ denotes the tokens processed per expert, each GPU processes a total of $2 \texp$ tokens across its local experts per MoE layer.

Assuming a uniform distribution of tokens, the all-to-all communication volume per GPU per MoE layer is given by:
\begin{equation*}
M_{\mathrm{comm}} = 2.5 \cdot \left( \frac{N}{\ep} \cdot \texp \cdot d \right) = 5 \cdot \texp \cdot d.
\end{equation*}
Here, the factor 2.5 accounts for the mixed-precision traffic (0.5 bytes for FP4 dispatch, 2 bytes for BF16 aggregation). On the compute side, the total FLOP count for the two local experts is:
\begin{equation*}
C_{\mathrm{comp}} = 2 \cdot \left( \frac{N}{\ep} \cdot \texp \cdot d \cdot m \right) = 4 \cdot \texp \cdot d \cdot m.
\end{equation*}
The corresponding compute time is: $t_{\mathrm{comp}} = \frac{C_{\mathrm{comp}}}{F} = \frac{4 \cdot \texp \cdot d \cdot m}{F}$. Similarly, the all-to-all communication time is:
$t_{\mathrm{comm}} = \frac{M_{\mathrm{comm}}}{\mathrm{BW}_{\mathrm{NVL}}} = \frac{5 \cdot \texp \cdot d}{\mathrm{BW}_{\mathrm{NVL}}}$,
where $\mathrm{BW}_{\mathrm{NVL}} = 900~\text{GB/s}$ is the effective unidirectional NVLink bandwidth. Consequently, the ratio of communication time to compute time is:
\begin{equation*}
\frac{t_{\mathrm{comm}}}{t_{\mathrm{comp}}} = \frac{5 \cdot \texp \cdot d / \mathrm{BW}_{\mathrm{NVL}}}{4 \cdot \texp \cdot d \cdot m / F} = \frac{5 \cdot F}{4 \cdot m \cdot \, \mathrm{BW}_{\mathrm{NVL}}}.
\end{equation*}
Substituting the parameters for the GB200 NVL72 and Qwen3-235B-A22B yields a ratio of approximately 9. This indicates that in the throughput-oriented regime, MoE layers are heavily dominated by all-to-all communication overhead.

\begin{notebox}{Design Principle II}
Improving performance in throughput-oriented MoE deployments requires minimizing the data volume of all-to-all operations. This volume is proportional to:
\begin{equation*}
M_{\mathrm{comm}} \propto \frac{N}{\ep} \cdot \texp \cdot d = \frac{\ttotal \cdot K \cdot d}{\ep}.
\end{equation*}
Consequently, communication overhead can be mitigated by reducing the routed hidden dimension $d$ or the number of active experts $K$. Note that modifying the intermediate dimension $m$ does not affect the token size and thus yields no direct improvement.
\end{notebox}

\subsection{Model Quality}
Beyond optimizing inference speed, preserving model quality is paramount. To this end, we draw on theoretical insights into neural network expressivity and combinatorial sparsity.
Classical results on Barron functions~\citep{barron1993} state that a one-hidden-layer network with $u$ nonlinear units achieves a mean-squared error of $\mathcal{O}(1/u)$, independent of the input dimension $d$. In an MoE layer, this effective nonlinear budget per token is proportional to the total width of the selected experts:
\begin{equation*}
U_{\mathrm{eff}} \propto K \cdot m.
\end{equation*}
This implies that reducing the active experts $K$ or the intermediate dimension $m$ directly penalizes the effective capacity ($U_{\mathrm{eff}}$), risking model quality degradation.

\begin{notebox}{Design Principle III}
Maintaining model quality requires preserving the effective nonlinear budget, $K \cdot m$. Consequently, to alleviate memory and communication bottlenecks without sacrificing model quality, we should keep both the number of active experts and the intermediate dimension unchanged.
\end{notebox}

Every inference task is characterized by an intrinsic feature rank, $r_\mathrm{eff}$, corresponding to the minimum number of degrees of freedom required to preserve task-relevant information. Reducing the hidden dimension $d$ below this threshold necessarily discards such information, leading to accuracy degradation. Thus, $r_\mathrm{eff}$ serves as a task-dependent lower bound on $d$.

\begin{notebox}{Design Principle IV}
There exists a task-specific feature rank $r_\mathrm{eff}$ that imposes a lower limit on the reduction of $d$. Reducing $d$ below this limit precipitates a collapse in model quality.
\end{notebox}

Additionally, the MoE architecture benefits from combinatorial sparsity, offering $\binom{N}{K}$ possible expert combinations per token~\citep{deepseekmoe}. Increasing the total number of experts $N$ expands this specialization space. Furthermore, scaling both $N$ and $K$ by a factor $\alpha$ exponentially increases the diversity of expert mixtures:
\begin{equation*}
    \binom{\alpha N}{\alpha K} \ge \left( \binom{N}{K} \right)^\alpha.
\end{equation*}

\begin{notebox}{Design Principle V}
Scaling both the number of experts $N$ and top-k per token $K$ enhances model quality by exponentially expanding the space of expert combinations.
\end{notebox}

\paragraph{Putting it all together.} Design Principles I and II indicate that improving inference speed requires reducing both memory bandwidth and communication costs. Memory bandwidth cost scales with 
$d$ and $m$, while communication cost scales with $K$ and $d$. However, Principle III cautions against reducing either $K$ or $m$, as doing so would likely degrade model quality. This leaves $d$ as the most promising dimension to reduce, enabling performance improvements in both throughput- and latency-oriented regimes without significant loss in accuracy. Principle IV further establishes a lower bound, ($r_\mathrm{eff}$), on $d$ to prevent quality collapse. Moreover, Principle V suggests that increasing $N$ and $K$ improves model quality. Since memory bandwidth and communication costs scale linearly with $K$, we can simultaneously increase $K$ by a factor $\alpha$ and reduce $d$ by the same factor $\alpha$ (provided $d/\alpha \ge r_\mathrm{eff}$). \textcolor{green!50!black}{\textit{We hypothesize, and empirically validate in subsequent sections, that this transformation (also depicted in Figure~\ref{fig:latent_moe}) preserves memory bandwidth and communication costs while improving network expressivity and combinatorial sparsity, yielding higher accuracy per FLOP and per parameter.}}

\section{LatentMoE Architecture} \label{sec:latentmoe}
Guided by the design principles outlined in Section~\ref{sec:design_choices}, we introduce LatentMoE, a new MoE architecture designed for efficient scaling. LatentMoE first projects each input token $x \in \R^d$ into a lower-dimensional latent space $\R^{\ell}$ using a learnable down-projection matrix $W_{\downarrow} \in \R^{\ell \times d}$. The resulting compressed representation is then routed to the selected experts. Each expert $E_i(\cdot; \ell)$ operates entirely within the latent space and is parameterized by weights $\wfconei, \wgatei \in \R^{m \times \ell}$ and $\wfctwoi \in \R^{\ell \times m}$. After expert computation, the outputs are aggregated and projected back to the original input dimension using a learnable up-projection matrix $W_\uparrow \in \R^{d \times \ell}$.

Since we compress only the input dimension $d$ to $\ell$ while keeping the intermediate dimension $m$ constant, the effective nonlinear budget $U_\mathrm{eff}$ remains unchanged. While Design Principle III suggests this should theoretically preserve accuracy, in practice, larger models are often easier to train and more robust to hyperparameter variations~\citep{lottery_ticket_hypothesis, novak2018sensitivity, sensitivity_analysis}. To avoid extensive hyperparameter tuning for the compressed model, we leverage Design Principle V by scaling the total number of experts $N$ by a factor $\alpha = d/\ell$, thereby expanding the combinatorial specialization space.
Crucially, since neither the memory bandwidth cost (in latency-oriented scenarios) nor the communication cost (in throughput-oriented scenarios) depends on $N$, this scaling adheres to Design Principles I and II, incurring no additional inference overhead. Hereafter, we refer to this architecture modification as $\lmoeeff$, formally defined as follows:
\begin{equation}
\label{eq:latent_moe_eff}
\lmoeeff(x) := W_{\uparrow} \cdot \left( \sum_{i \in \mathcal{T}_{K,N'}} p'_i E_i(W_\downarrow \cdot x; \ell) \right) + \sum_{j=N'+1}^{N'+S} E_j(x; d).
\end{equation}
Here, $N' = \alpha \cdot N$ denotes the expanded set of routed experts. The routed experts $E_i(\cdot; \ell)$ function within the latent space, while the shared experts $E_j(\cdot; d)$ operate in the original input space. The routing weights $p' = \softmax(W'_r \cdot x)$ are computed from the original token $x \in \R^d$ using a learnable weight matrix $W'_r \in \R^{N' \times d}$, and $\mathcal{T}_{K,N'}$ denotes the indices of the top-$K$ experts (out of $N'$ total) selected based on their routing scores. For simplicity, all operations outside the routed experts---including the MoE routing mechanism and shared experts---continue to operate in the original hidden dimension $d$, as they do not significantly contribute to the identified memory and communication bottlenecks.

Following the down-projection $W_\downarrow$, token dispatch and aggregation occur in the latent space $\R^\ell$. This reduces the communication volume by a factor of $\alpha$ relative to a standard MoE. Similarly, because the expert weights lie in the latent space ($\R^{m \times \ell}$ and $\R^{\ell \times m}$), the memory bandwidth cost for weight loading is also reduced by a factor of $\alpha$.

Design Principle V further suggests that scaling both $N$ and $K$ by a factor $\alpha$ exponentially increases expert diversity, thereby enhancing model quality. Following this principle, the default LatentMoE configuration (a.k.a., $\lmoeacc$) is defined as follows: 
\begin{equation}
\label{eq:latent_moe_acc}
\lmoeacc(x) := W_{\uparrow} \cdot \left( \sum_{i \in \mathcal{T}_{K',N'}} p'_i E_i(W_\downarrow \cdot x; \ell) \right) + \sum_{j=N'+1}^{N'+S} E_j(x; d),
\end{equation}
where $K' = \alpha \cdot K$. This formulation differs from $\lmoeeff$ solely in the number of active experts, utilizing the top-k selection function $\mathcal{T}_{K',N'}$.

Since $K$ is increased by a factor of $\alpha = d / \ell$, this variant keeps communication cost and memory bandwidth requirements constant relative to a standard MoE. The increased expert diversity and non-linearity budget per token, however, lead to superior model accuracy at iso-inference cost, thereby pushing the Pareto frontier of models to a new level. Table~\ref{tbl:moe_costs} summarizes the costs and benefits of the two configurations, $\lmoeeff$ and $\lmoeacc$. For completeness, we evaluate both setups in Section~\ref{sec:results}.

\begin{table}[t]
\centering
\caption{Comparison of asymptotic communication and memory bandwidth costs per GPU. Costs are normalized by hardware constants. $T_{\text{exp}}$ denotes the average number of tokens per expert, and $\ep$ is the expert-parallel level. Arrows indicate improvement ($\uparrow$), maintenance ($\rightarrow$), or baseline (--).}
\label{tbl:moe_costs}
\begin{tabular}{lcccc}
\toprule
\textbf{Architecture} & \textbf{\makecell{Communication\\Cost}} & \textbf{\makecell{Weight Loading\\Memory Cost\\per Expert}} & \textbf{\makecell{Model\\Accuracy}} & \textbf{\makecell{Inference\\Efficiency}} \\
\midrule
Standard MoE
  & $(N / \ep) \cdot \, \texp \cdot d$
  & $d \cdot m$
  & --
  & -- \\
\addlinespace
$\lmoeeff$
  & $(N / \ep) \cdot \, \texp \cdot \ell$
  & $\ell \cdot m$
  & $\rightarrow$
  & \textcolor{green!50!black}{$\uparrow$} \\
\addlinespace
$\lmoeacc$ (recommended)
  & $(N / \ep) \cdot \, \texp \cdot d$
  & $d \cdot m$
  & \textcolor{green!50!black}{$\uparrow$}
  & $\rightarrow$ \\
\bottomrule
\end{tabular}
\end{table}

\section{Evaluation}
\label{sec:results}
In this section, we conduct a thorough design space exploration to verify the effectiveness of the proposed LatentMoE architecture. We start by pretraining Transformer MoE models at two different scales: (1) 16B total parameters with 2B active, which we use for conducting ablation studies, and (2) 95B total parameters with 8B active, which we use as a scaling test of the 16B results. To demonstrate the generalizability of LatentMoE architectures, we further extend our study by training hybrid Mamba-Attention MoE models at scale. 

We use the architecture and hyperparameters from DeepSeek-v2-lite~\citep{deepseek_v2} for our 2B active model ablations. For the 8B active Transformer model, we use a cosine learning rate schedule with a max learning rate of $1.2 \times 10^{-3}$ decayed to a minimum of $3 \times 10^{-6}$. The 8B active Hybrid model is trained with a WSD schedule with a max learning rate of $8 \times 10^{-4}$ decayed to $8 \times 10^{-6}$ in the last 15\% of training.  Both the 8B active Transformer and hybrid models are trained with a sequence length of 8192, a batch size of 768 ($\sim6$ million tokens), and a learning rate warmup of $8.4$ billion tokens. The 8B active models use a load balancing loss coefficient of $10^{-4}$ along with DeepSeek’s aux-loss-free load balancing strategy~\citep{auxiliarylossfreeloadbalancingstrategy} to ensure balanced token load throughout training. Table~\ref{tbl:model_arch} summarizes the model architecture under study in this paper. 

\begin{table}[ht!]
    \centering
    \caption{Architectural specifications of the baseline models used for design space exploration. For the Hybrid model's Mamba layers, we use Mamba-2 blocks with 128 heads, 64 head dimension, 128 state dimension, and 8 groups.}
    \label{tbl:model_arch}
    \resizebox{\textwidth}{!}{%
    \begin{tabular}{lccc}
    \toprule
    \textbf{Configuration} & \textbf{16BT-2BA} & \textbf{95BT-8BA} & \textbf{Hybrid-73BT-8BA} \\
    \midrule
    Layers ($L$) & 27 & 32 & 52 (24 Mamba/MoE, 4 Attn.)\\
    Hidden Dimension ($d$) & 2048 & 4096 & 4096\\
    Total Routed Experts ($N$) & 64 & 128 & 128 \\
    Active Experts ($K$) & 6 & 6 & 6 \\
    Shared Experts ($S$) & 2 & 2 & 2 \\
    Intermediate FFN Dimension ($m$) & 1408 & 2688 & 2688 \\
    Activation Function & SwiGLU & Squared-ReLU & Squared-ReLU \\
    Attention Heads & 16 & 32 & 32 \\
    Query Groups (GQA) & 16 & 8 & 8 \\
    \midrule
    Total Parameters & 16B & 95B & 73B \\
    Active Parameters & 2B & 8B & 8B \\
    \bottomrule
    \end{tabular}
    }
\end{table}

\subsection{LatentMoE Ablations}

\begin{figure}[ht!]
    \centering
    \includegraphics[width=0.7\textwidth]{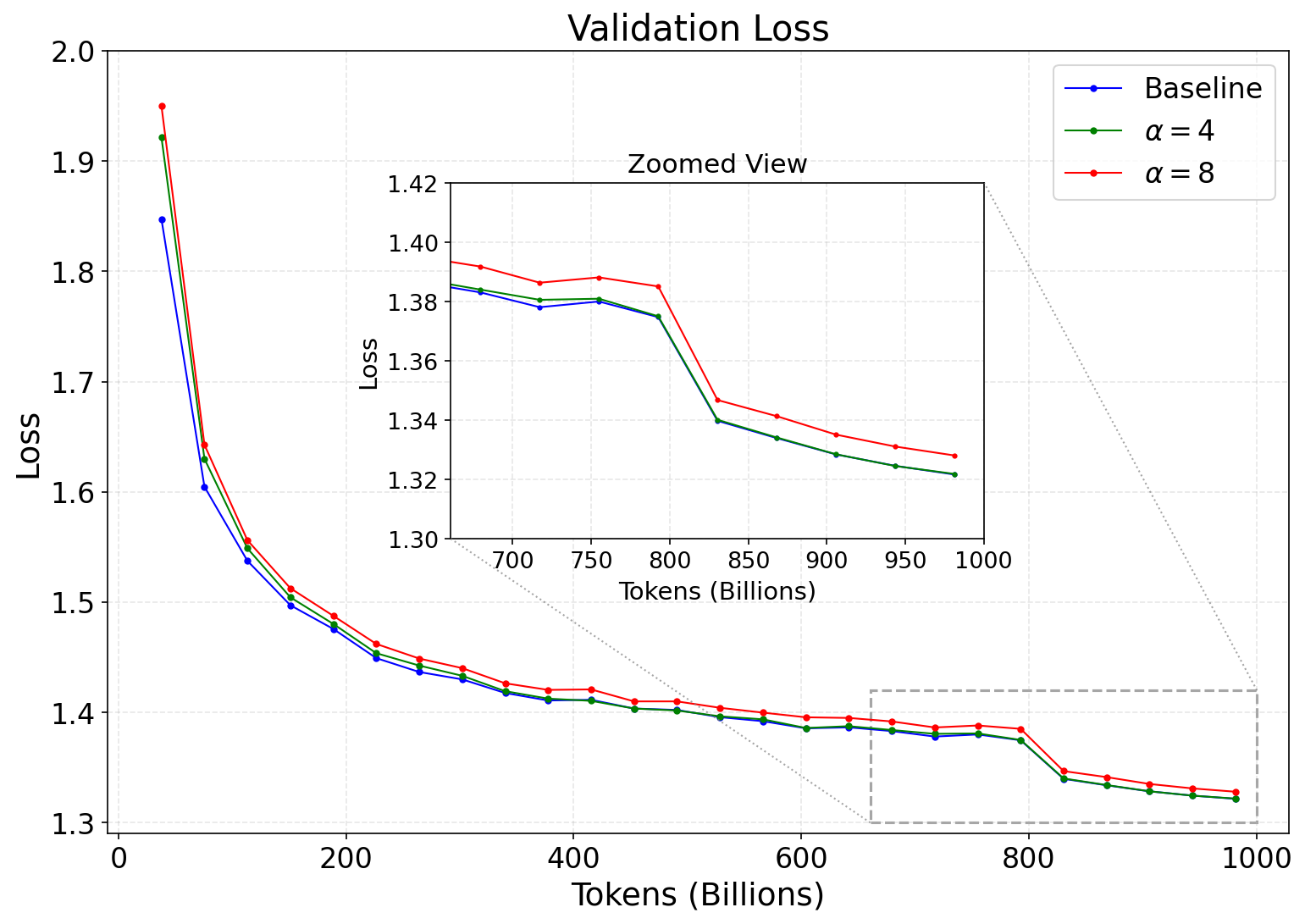}
    \caption{\textbf{Effect of compression ratio on model quality.} Validation loss for the 16BT-2BA model using the $\lmoeeff$ configuration across varying compression ratios $\alpha=d/\ell$. The total number of experts is scaled by $\alpha$, while the base model configuration follows Table~\ref{tbl:model_arch}.}
    \label{fig:results_2b_vary_compression}
\end{figure}

\newparagraph{Impact of compression ratio.} Design Principle IV (Section~\ref{sec:design_choices}) hypothesizes that there exists an intrinsic rank $r_{\mathrm{eff}}$ such that compressing the latent dimension to $\ell \ge r_{\mathrm{eff}}$ results in negligible information loss. To empirically validate this and estimate $r_{\mathrm{eff}}$, we pretrain and sweep different compression ratios on top of the $\lmoeeff$ configuration, holding all other hyperparameters constant.
Results in Figure~\ref{fig:results_2b_vary_compression} indicate that model quality is preserved for compression ratios $\alpha \le 4$. Consequently, we adopt $\alpha = 4$ for all subsequent experiments. We empirically verified that this setting remains effective at larger scales as well (i.e., 95B total and 8B active).

\begin{figure}[ht!]
    \centering
    \includegraphics[width=0.7\textwidth]{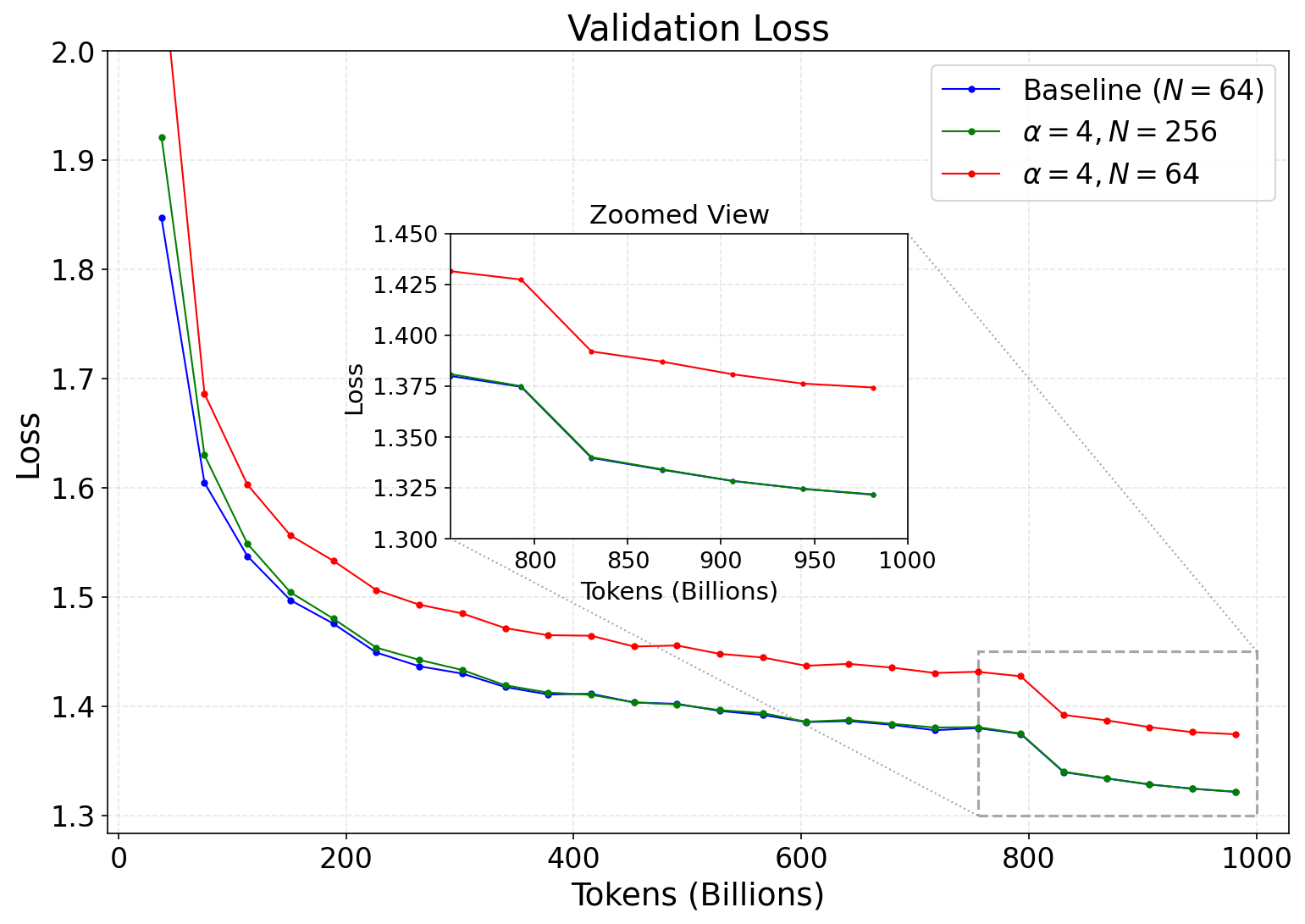}
    \caption{\textbf{Impact of expert scaling on model quality.} Comparison of validation loss for the 16BT-2BA model using the $\lmoeeff$ configuration when the hidden dimension is compressed by $4\times$. The green curve utilizes the proposed expert scaling ($N' = \alpha N$), while the red curve does not. Scaling the expert count effectively mitigates the accuracy loss caused by compression, eliminating the need for extensive hyperparameter retuning.}
    \label{fig:results_reduced_params}
\end{figure}

\newparagraph{Impact of number of experts.}
In Section~\ref{sec:latentmoe}, we noted that parameter reduction via compression can impede training stability. To quantify this, we pretrain the $\lmoeeff$ LatentMoE variant of the 16B total and 2B active parameter model with the hidden dimension $d$ compressed by a factor of 4, both with and without a compensatory increase in the total number of experts, using the baseline hyperparameters. As shown in Figure~\ref{fig:results_reduced_params}, reducing $d$ without scaling the expert count leads to significant quality degradation, validating the expert scaling strategy employed by LatentMoE.

\begin{figure}[h]
\centering
\includegraphics[width=0.7\textwidth]{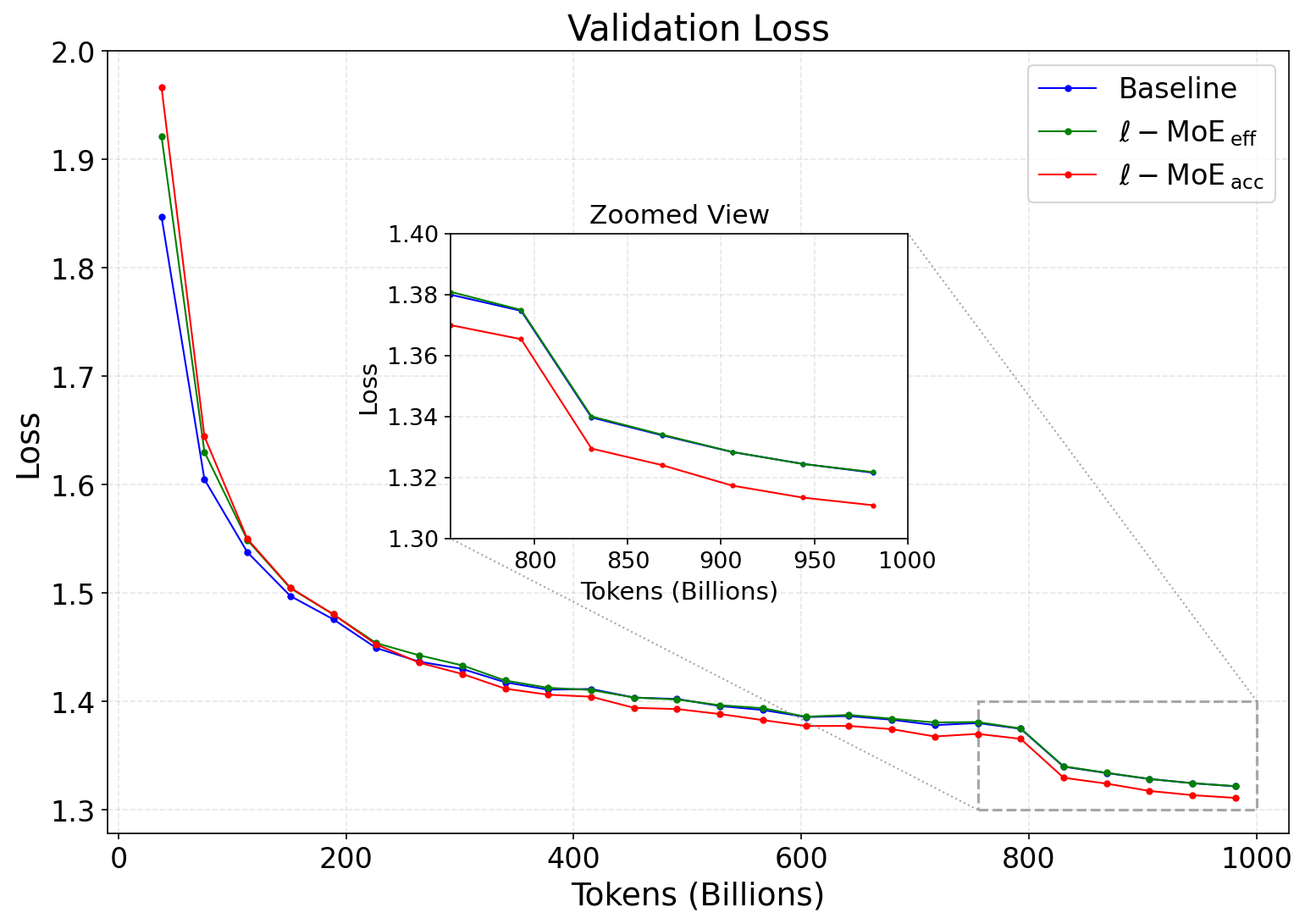}
\caption{\textbf{Comparison between LatentMoE variants.} Training trajectories for the baseline 16BT-2BA
model versus the $\lmoeeff$ and $\lmoeacc$ ($\ell=512$). $\lmoeeff$ matches baseline convergence, while $\lmoeacc$ outperforms the baseline. }
\label{fig:results_2b_latent512_baseline_leff_lacc}
\end{figure}

\newparagraph{Comparison between the two variants of LatentMoE.}
In Section~\ref{sec:latentmoe}, we introduced two LatentMoE variants: $\lmoeeff$, designed to improve inference efficiency while maintaining baseline accuracy, and $\lmoeacc$, designed to enhance accuracy at a comparable inference cost. Figure~\ref{fig:results_2b_latent512_baseline_leff_lacc} compares the validation loss of these configurations against the baseline for the 16B total and 2B active parameter model using a latent dimension of $\ell=512$ ($\alpha=4$). Consistent with our expectations, $\lmoeeff$ matches the baseline accuracy, whereas $\lmoeacc$ achieves a noticeably lower validation loss. \textcolor{green!50!black}{\textit{We recommend $\lmoeacc$ for Pareto-optimal accuracy versus inference cost.}} \\

\subsection{LatentMoE Scaling Studies}
\label{sec:accuracies}
\begin{figure}[ht!]
\centering
\includegraphics[width=0.7\textwidth]{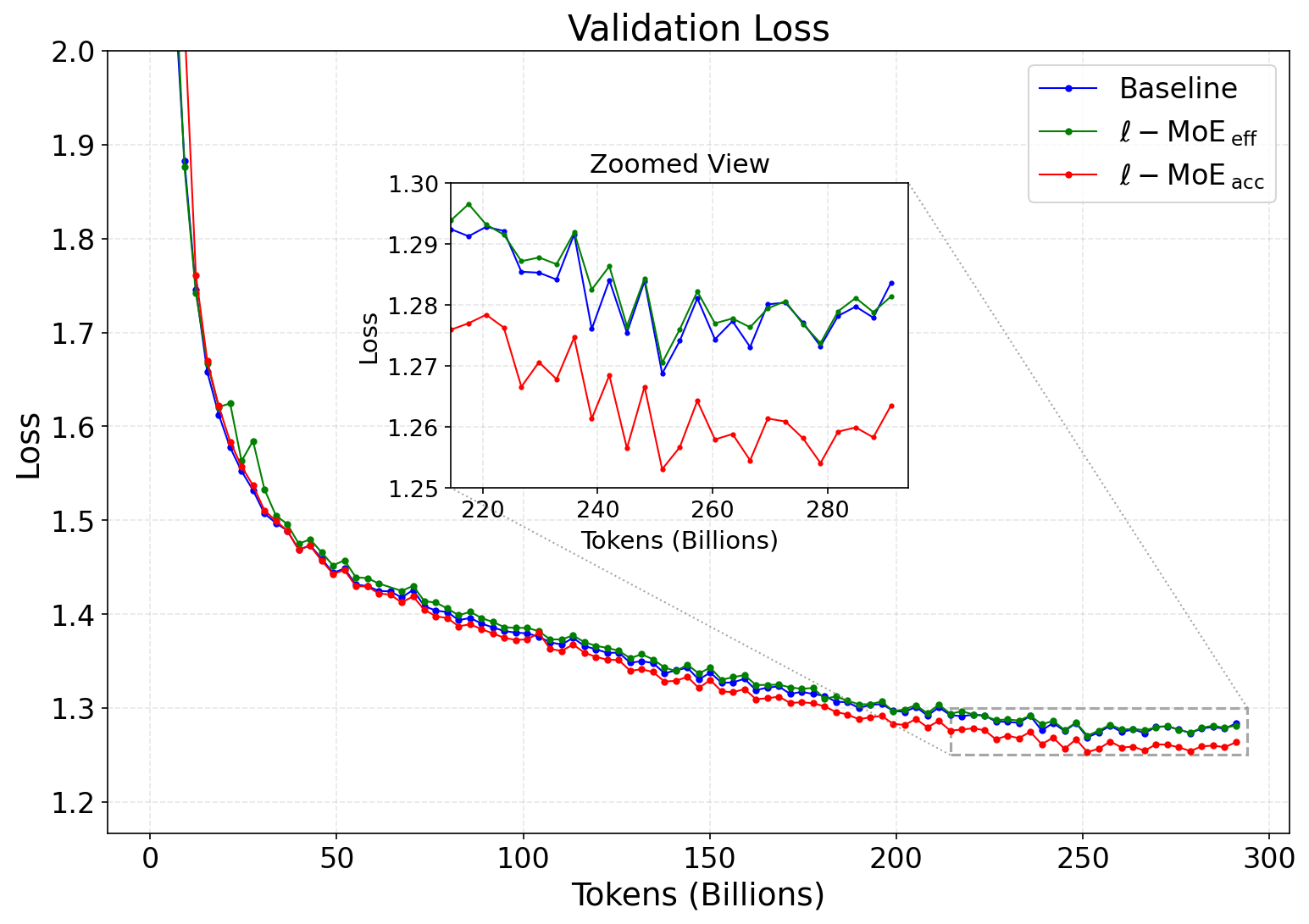}
\caption{\textbf{95B Model Training Convergence.} Validation loss curves for the 95BT-8BA baseline, $\lmoeeff$, and $\lmoeacc$ configurations ($\ell=1024, \alpha=4$). $\lmoeeff$ matches baseline convergence, while $\lmoeacc$ outperforms the baseline.}
\label{fig:8b_val_loss}
\end{figure}

Leveraging the insights from the 16B model ablations, we train a 95B parameter Transformer using a LatentMoE configuration with a $4\times$ compression ratio. Figure~\ref{fig:8b_val_loss} presents the validation loss trajectories for $\lmoeeff$ and $\lmoeacc$ relative to the baseline. Consistent with the 16BT-2BA results, $\lmoeeff$ matches the baseline, while $\lmoeacc$ demonstrates superior results. Table~\ref{tbl:8b_acc} shows the downstream task accuracy at the 300B token horizon.
We report Code as the average over HumanEval, HumanEval+, MBPP, and MBPP+, Math as the average of GSM8K CoT and MATH-500, and Commonsense understanding as the average of RACE, ARC-Challenge, HellaSwag, and Winogrande.
For simplicity, we used the exact same hyperparameters optimized for the baseline Transformer for LatentMoE. Further hyperparameter tuning might lead to even better accuracy.

\begin{table}[ht!]
    \centering
    \caption{Accuracy comparisons of the 95BT-8BA model with and without LatentMoE. Compared to the baseline, with equivalent parameters, $\lmoeacc$ provides higher accuracy across all downstream tasks, while $\lmoeeff$ provides comparable to or better accuracy with much fewer FLOPs.}
    \label{tbl:8b_acc}
    \resizebox{\textwidth}{!}{%
    \begin{tabular}{lccccccc}
    \toprule
    \textbf{Model} & \textbf{Active Params} & \textbf{Total Params} & \textbf{MMLU Pro} & \textbf{MMLU} & \textbf{Code} & \textbf{Math} & \textbf{Commonsense} \\
    \midrule
    Baseline & 8.47B & 94.4B & 29.26 & 58.95 & 40.33 & 64.39 & 74.32 \\
    \midrule
    $\lmoeacc$ & 8.44B & 94.8B & \textbf{34.91} & \textbf{62.23} & \textbf{41.50} & \textbf{64.88} & \textbf{75.18} \\
    $\lmoeeff$ & 5.62B & 94.8B & 34.75 & 61.06 & 40.68 & 63.61 & 73.72 \\
    \bottomrule
    \end{tabular}
    }
\end{table}

To further validate the effectiveness of the LatentMoE architecture, we also pretrain a series of hybrid Mamba-Attention MoE models. Specifically, we first train a baseline 8B active (73B total) parameter model. As described in Table~\ref{tbl:model_arch}, each MoE layer in the hybrid architecture contains 128 experts, 6 activated experts, 2 shared experts, and uses an intermediate FFN dimension of 2688. We use Squared-ReLU activation and a model dimension of 4096. We then train the $\lmoeeff$ and $\lmoeacc$ LatentMoE variants of the baseline model, using a $4\times$ compression ratio.

Results after training the above models on 1T tokens are shown in Table~\ref{tbl:hybrid_8b_1t}. All models are trained with identical hyperparameters. As shown, the LatentMoE $\lmoeacc$ variant achieves significantly higher accuracy than the baseline across all tasks, while the $\lmoeeff$  variant achieves accuracy comparable to or better than the standard granular MoE baseline. 

Overall, the LatentMoE architecture provides a clear advantage in terms of accuracy per FLOP and per parameter compared to granular MoEs, paving the way for higher accuracy at fixed inference cost or lower inference cost at fixed accuracy.

\begin{table}[h]
    \centering
    \caption{Accuracy comparisons of hybrid Mamba-Attention MoEs with and without LatentMoE. Compared to the baseline, with equivalent parameters, $\lmoeacc$ provides higher accuracy across all downstream tasks, while $\lmoeeff$ provides comparable or better accuracy with much fewer FLOPs.}
    \label{tbl:hybrid_8b_1t}
    \resizebox{\textwidth}{!}{%
    \begin{tabular}{lccccccc}
    \toprule
    \textbf{Model} & \textbf{Active Params} & \textbf{Total Params} & \textbf{MMLU Pro} & \textbf{MMLU} & \textbf{Code} & \textbf{Math} & \textbf{Commonsense} \\
    \midrule
    Baseline & 8.09B & 72.6B & 48.30 & 70.10 & 51.95 & 78.32 & 81.73 \\
    \midrule
    $\lmoeacc$ & 8.02B & 72.8B & \textbf{52.87} & \textbf{72.11} & \textbf{55.14} & \textbf{80.19} & \textbf{82.10} \\
    $\lmoeeff$ & 5.91B & 72.8B & 51.29 & 71.34 & 53.13 & 77.01 & 80.78 \\
    \bottomrule
    \end{tabular}
    }
\end{table}

\subsection{Inference Performance}
As discussed in Section~\ref{sec:latentmoe}, $\lmoeacc$ is expected to achieve similar inference speed to the standard MoE baseline while attaining higher accuracy.
Our evaluations in Section~\ref{sec:accuracies} confirmed that $\lmoeacc$ indeed achieves higher accuracy compared to standard MoE. Here, we evaluate $\lmoeacc$ from the perspective of inference efficiency.
Table~\ref{tbl:measuredperf} presents the measured performance of $\lmoeacc$ compared to standard MoE for the Hybrid-73BT-8BA model (see Table~\ref{tbl:model_arch}) on two Hopper H100 GPUs using vLLM with FP8 per-tensor quantization.
We focus our measurements on the hybrid Mamba-Attention baseline, as it represents the most efficient inference architecture.

\begin{table}[ht!]
    \centering
    \footnotesize
    \begin{tabular}{ccc}
    \hline
     \textbf{Concurrency} & \textbf{LatentMoE (Tokens/s/GPU)} & \textbf{Standard MoE (Tokens/s/GPU)} \\
    \hline
   1 & 181.6 & 206.6 \\
     4 & 528.5 & 509.8 \\
    16 & 1130.8 & 1204.6 \\
     64 & 1569.6 & 1549.3 \\
     128 & 1625.8 & 1725.9 \\
    \hline
    \end{tabular}
    \caption{Comparison of LatentMoE and Standard MoE performance metrics.}
    \label{tbl:measuredperf}
\end{table}

The measurements show that at higher concurrencies, per-GPU throughput drops by only up to 6\%. It is important to note that further software optimizations could be performed to mitigate even these small throughput differences between LatentMoE and standard MoE. One proposed optimization is to utilize separate CUDA streams for routed and shared experts, which could reduce end-to-end latency when performing inference with smaller batches or when model dimensions do not saturate the GPU's compute. A second optimization targets the MoE kernels from the CUTLASS library. Since LatentMoEs decrease the size of the GEMMs for routed experts, it is important to ensure that inner dimensions remain large enough to fully utilize the GPU and avoid SM-bound workloads. When inner dimensions are small, specialized smaller-matrix GEMM kernels should be used.

\subsubsection{Projected Serving Impact at Trillion-Parameter Scale}
\label{sec:discussion_projected_serving} 

Inference efficiency can be characterized as a three-dimensional trade-off surface, with accuracy along one axis, throughput per GPU along a second axis, and latency (user interactivity) along the third axis. Thus far, we have discussed the accuracy of LatentMoE and presented measured performance at the 95B-parameter scale. In the following section, we examine a two-dimensional slice of this trade-off at the trillion-parameter scale by projecting throughput per GPU and latency Pareto frontiers for accuracy-matched models.

\begin{figure}[ht!]
  \centering
  \includegraphics[width=1.0\linewidth]{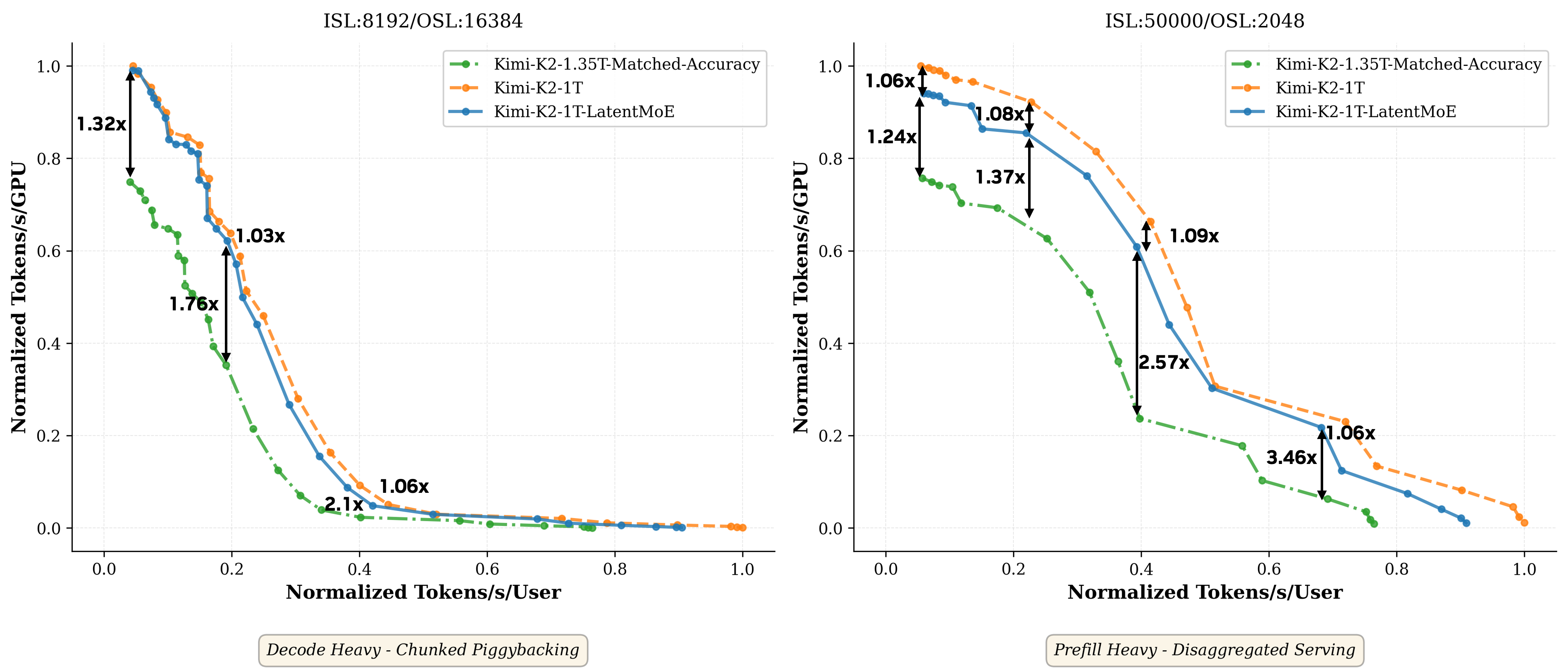}
  \caption{\textbf{Projected throughput-latency Pareto frontiers at trillion-parameter scale.}
  \textit{Left:} Decode-heavy traffic modeled with chunked piggybacking serving. \textit{Right:} Prefill-heavy traffic modeled with disaggregated serving.
  \textsc{Kimi-K2-1T-LatentMoE} attains accuracy-efficient operating points. Matching its accuracy under standard MoE scaling requires an additional $\sim$350B parameters in our construction and yields a $1.24\times$--$3.46\times$ projected slowdown across the frontier.}
  \label{fig:normalized_plot}
\end{figure}

\noindent \textbf{Performance evaluation methodology.} We use a high-fidelity proprietary performance simulator to project end-to-end serving performance for a trillion-parameter class model and its LatentMoE variant. We simulate over $200$K operating points to estimate the throughput per GPU and latency Pareto frontiers shown in Figure~\ref{fig:normalized_plot}. We consider two traffic patterns. The first is a decode-heavy setting modeled with chunked piggybacking serving. The second is a prefill-heavy setting modeled with disaggregated serving, where prefill and decode are separated to reflect long-context traffic. The serving strategy (for example, whether to use disaggregation) is selected following the guidance in~\cite{mitra2025beyond}.

\noindent \textbf{Effective Parameter Multiplier (EPM).} We use the effective parameter multiplier to construct an iso-accuracy baseline for inference comparisons. We benchmark the native \textsc{Kimi-K2-1T} against our proposed variant, \textsc{Kimi-K2-1T-LatentMoE}. Following the ``Effective Parameter Count'' framework established by~\cite{frantar2025compressionscaling}, we posit that a treated model with physical parameters $N_{treat}$ behaves like a standard dense baseline with effective parameters $N_{eff}$. We assume baseline performance follows a scaling law $f(N)$. For a treated model achieving score $S_{treat}$, the effective capacity is obtained by inverting the baseline scaling law:
\begin{equation}
    N_{eff} = f^{-1}(S_{treat})
\end{equation}
The EPM is defined as the ratio of effective capacity to physical parameters:
\begin{equation}
    \lambda = \frac{N_{eff}}{N_{treat}}
\end{equation}
We use $\lambda$ to construct an iso-accuracy baseline with parameter count $N_{\text{iso}} = \lambda \cdot N_{treat}$. For our evaluations, we derive $f(N)$ by fitting a log-linear function to the MMLU accuracy scores of the \textsc{Qwen-3-Dense} model family (0.6B, 1.7B, 4B, 8B, 14B, and 32B):
\begin{equation}
    f(N) = a \cdot \log N + b
\end{equation}
where $a$ and $b$ are fitted parameters.

\noindent \textbf{Iso-accuracy baseline construction.} We estimate an Effective Parameter Multiplier of $\lambda \approx 1.35\times$ for \textsc{Kimi-K2-1T-LatentMoE}. For a 1T-parameter base model, this implies an iso-accuracy scale of $N_{\text{iso}} \approx 1.35$T, corresponding to an increase of $(1.35 - 1.0)$T $\approx 0.35$T $\approx 350$B parameters. Guided by this target, we construct a physical \textbf{iso-accuracy baseline}, denoted \textsc{Kimi-K2-1.35T}, by scaling the native architecture depth from 61 to 80 layers. This construction matches the projected effective capacity implied by LatentMoE and enables a direct inference-efficiency comparison against a standard model of comparable predictive power.

\noindent \textbf{Accuracy matching with standard MoE scaling is more expensive.} $\lmoeacc$ achieves an accuracy gain at fixed parameter and FLOP budget. When we enforce an accuracy-matched comparison using the standard MoE architecture, the required scaling incurs a marked serving penalty. Across the projected Pareto frontier (Figure~\ref{fig:normalized_plot}), \textsc{Kimi-K2-1.35T} is approximately $1.24\times$--$3.46\times$ slower than \textsc{Kimi-K2-1T-LatentMoE}, indicating that $\lmoeacc$ provides a more favorable accuracy-latency trade-off than increasing model size through the standard MoE architecture to reach the same accuracy target.

\noindent \textbf{Latent projection overhead is modest.} Relative to native \textsc{Kimi-K2-1T}, \textsc{Kimi-K2-1T-LatentMoE} introduces additional computation due to latent projection operators. In our projections, native \textsc{Kimi-K2-1T} remains close, within up to $\sim$9\% of \textsc{Kimi-K2-1T-LatentMoE}, indicating that projection overhead is small compared to the cost of achieving the same accuracy gain via standard scaling to \textsc{Kimi-K2-1.35T}.

\section{Related Work}
\label{sec:discussion}

Mixture-of-Experts (MoE) models have become a cornerstone of state-of-the-art large language model services. In this work, we challenge the original MoE design paradigm for the first time and introduce an alternative architecture that achieves higher accuracy under both iso-parameter and iso-FLOP constraints.

In parallel, the community has developed a rich set of model compression techniques to reduce inference cost, including quantization~\citep{rouhani2023microscalingdataformatsdeep, shared_microexponents} and sparsity~\citep{xie2024moeprunerpruningmixtureofexpertslarge}. At the expert level, pruning~\citep{not_all_experts_are_equal, reap_the_experts, chen2022taskspecificexpertpruningsparse} and merging~\citep{li2024mergecompressdemystifyefficient} methods have also been proposed. These approaches are orthogonal to the LatentMoE design and can be composed with it to yield further efficiency gains.

The closest related work to this paper is probably MoLAE~\citep{molae}. MoLAE is a post-training compression method built on low-rank approximation of expert weights in a latent space. Although the two methods appear similar at a surface level, LatentMoE makes fundamentally different design trade-offs by coupling expert compression with increased network expressivity and combinatorial sparsity. By contrast, to compensate for accuracy loss caused by latent-space projection, MoLAE introduces grouped latent projections and restricts compression to only part of the experts (FC2). These design choices, in turn, forgo communication savings during token dispatch and limit memory bandwidth reduction, ultimately constraining the achievable efficiency gains. As discussed in Section~\ref{sec:design_choices}, efficient MoE serving is not FLOP-bound; reducing FLOPs alone is not enough to improve the Pareto frontier of accuracy vs. throughput vs. latency.

Concurrent work explores improving model quality under fixed compute by modifying residual connectivity rather than the expert path. Manifold-Constrained Hyper-Connections (mHC)~\citep{xie2026mhcmanifoldconstrainedhyperconnections} improves quality under iso-compute by widening the residual stream and increasing residual-path connectivity.
Achieving this requires a materially different residual topology (multi-stream residual state) and a learned connection-generation mechanism (RMSNorm $\rightarrow$ linear $\rightarrow \tanh$ gating for the connection maps, plus constrained residual mixing for stability). We believe LatentMoE and mHC are complementary and can be stacked on top of one another. Further exploration is left to future work.

\section{Conclusion}
\label{sec:conclusion}
We presented LatentMoE, a revised Mixture-of-Experts architecture designed to maximize accuracy per FLOP and per parameter by explicitly accounting for the dominant memory bandwidth and communication bottlenecks in modern inference systems. By projecting tokens into a lower-dimensional latent space, LatentMoE reduces routing all-to-all communication as well as the memory bandwidth and computation required per expert. These savings are then reinvested into scaling expert count and routing diversity without increasing inference cost. Across extensive experiments up to 95B parameters, hybrid architectures, and projected trillion-parameter serving scenarios, LatentMoE consistently outperforms standard MoEs on the accuracy–efficiency Pareto frontier.

\bibliography{references}
\bibliographystyle{references}

\end{document}